\documentclass[10pt,twocolumn,letterpaper]{article}

\usepackage{cvpr}              

\usepackage{svg}
\usepackage{kotex}
\usepackage{multirow}   
\usepackage{xcolor}
\definecolor{mygreen}{HTML}{2CA02C}
\definecolor{myorange}{HTML}{D55E00}
\definecolor{myblue}{HTML}{1F77B4}

\definecolor{cvprblue}{rgb}{0.21,0.49,0.74}
\usepackage[pagebackref,breaklinks,colorlinks,allcolors=cvprblue]{hyperref}

\def\paperID{*****} 
\def\confName{CVPR}
\def\confYear{2026}

\title{ReSS: Residual-Restoring Sparse Attention for 3D Vision Transformers}

\author{
Yongsung Kim$^{1}$~~Jaehoon Lee$^{1}$~~Minjun Park$^{1  }$~~Wooseok Song$^{2}$~~Hun Hwangbo$^{1}$~~Sungroh Yoon$^{1,2,3}$\textsuperscript{†}\\
$^1$IPAI, $^2$ECE, $^3$AIIS, ASRI, INMC, ISRC\\
Seoul National University\\
{\tt\small \{libary753, jhcaptain7, minjunpark, cody1129, genchiprofac, sryoon\}@snu.ac.kr}
}

\begin{document}
\maketitle
\let\thefootnote\relax\footnotetext{†Corresponding author}
\begin{abstract}
3D vision transformers such as VGGT predict camera poses and scene geometry from multi-view images in a single forward pass, but their global attention over all concatenated view tokens dominates computation as the number of views grows.
To reduce this cost, SparseVGGT and HeSS sparsify attention at the block level, and both retain blocks with high attention probability.
However, we observe that attention probability poorly predicts how much the model's behavior actually changes when a block is removed, and we show that this mismatch is why performance collapses as sparsity increases.
In this paper, we propose ReSS (ReSidual-ReStoring Sparse Attention), which recasts block selection from a problem of maximizing the retained attention mass to one of minimizing the drift that sparsification leaves in the residual stream.
We introduce a drift score that quantifies how much each block shifts the residual, and, since the drift of a drop set depends on the directions of the contribution vectors rather than on their magnitudes alone, an iterative residual restoration procedure that refines the drop set as a whole.
Across three backbones and five datasets, ReSS preserves dense performance better than prior methods at matched sparsity. Two further results support drift as the quantity that governs the cost of sparsification: maximizing drift degrades performance faster than random selection, and plotted against realized drift instead of sparsity, all methods fall approximately onto a single curve.
Code is available at \href{https://github.com/libary753/ReSS}{https://github.com/libary753/ReSS}.
\end{abstract}    
\section{Introduction}
\label{sec:intro}
\begin{figure}[t]
    \centering
    \includegraphics[width=\linewidth]{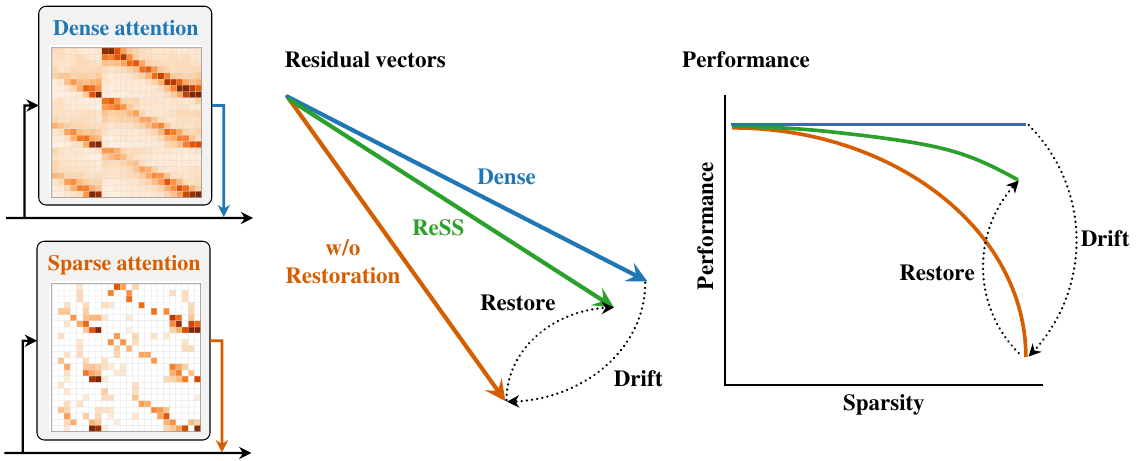}
    \caption{\textbf{Sparse selection by residual restoration.} Attention writes its output into the residual stream, and sparsification shifts what it writes; that shift is the residual drift. ReSS restores it by selecting the block set that minimizes this drift.}
    \label{fig:teaser}
\end{figure}

3D vision transformers (3D ViTs) such as VGGT~\cite{wang2025vggt}, $\pi^3$~\cite{wang2026pi}, and DepthAnything3~\cite{lin2025depthanything3} predict camera poses and scene geometry from multi-view images in a single forward pass. This capability comes from global attention over the concatenated tokens of all views. For the same reason, however, the attention cost grows quadratically with the number of views and dominates the overall computation precisely when many views are processed at once.

A common remedy is sparse attention: only a small fraction of query--key pairs contribute meaningfully to the output, so the rest can be skipped with little change; for practical speedups, the skipping is done at the block level. The question is then how to identify the blocks to keep. For 3D ViTs, SparseVGGT~\cite{wang2026sparseVGGT} first applies block-sparse attention on a pooled block-level attention map, and HeSS~\cite{kim2026hess} adds head-wise budget allocation. Both select them based on attention-probability magnitude, and both degrade rapidly as sparsity increases.

We trace this degradation to probability-based selection itself. When we directly measure how much the attention output changes upon removing a block, attention probability turns out to be a poor predictor of the ranking. The rank correlation fluctuates widely across layers and heads, dropping to near-random in later layers (\cref{sec:motivation}). In other words, existing methods decide what to keep without looking at the very quantity they aim to preserve: what attention writes into the residual stream. We call the shift that sparsification leaves there the residual drift (Fig. 1).

We propose \textit{ReSS} (\textbf{ReS}idual-\textbf{ReS}toring \textbf{S}parse Attention). ReSS recasts block selection as a set selection problem: find the block set that minimizes this drift. We introduce a drift score that quantifies how much removing each block shifts the residual. However, the drift of a drop set depends on the directions of the contribution vectors, not only on their magnitudes, so we pair the score with an iterative residual restoration procedure that refines the drop set as a whole. Each restoration round is fused into a single kernel to suppress its overhead.

Across three backbones and five datasets, and against baselines that include selection criteria ported from long-context LLM prefill and video diffusion, ReSS preserves dense performance best at matched sparsity. Moreover, maximizing drift by flipping the sign of the objective collapses performance faster than random selection. Re-plotted against the realized drift of each method rather than sparsity, performance falls approximately onto a single curve across methods. 

Our contributions are as follows.
\begin{itemize}
    \item We show that attention probability poorly predicts the output change caused by block removal, and recast block selection as minimizing residual drift over block sets.
    \item We introduce a drift score for single-block removal and iterative residual restoration that reduces the cumulative drift of the drop set.
    \item Across three 3D ViT backbones and five datasets, our method best preserves dense performance, in comparisons that include sparse attention methods designed for LLMs and video diffusion as well as for 3D ViTs.
    \item A sign-flipped control experiment and the collapse of performance onto a single curve as a function of drift support drift as the quantity that determines the cost of sparsification.
\end{itemize}
\section{Related Work}
\label{sec:related_work}

\paragraph{Accelerating 3D Vision Transformers}
Recent 3D ViTs replace multi-stage geometry pipelines~\cite{schonberger2016colmap} with a single feed-forward transformer that regresses camera poses and scene geometry from multi-view images~\cite{wang2024dust3r, wang2025vggt, wang2026pi, lin2025depthanything3}, at the cost of global attention over the concatenated tokens of all views. Token merging and pruning~\cite{bolya2023tome, shen2026fastvggt, shu2026litevggt, chen2026come} reduce the number of tokens entering attention, and architectural approaches replace attention with linear-time recurrence~\cite{zhuo2026streamvggt, chen2026ttt3r} or chunked processing~\cite{deng2025vggtlong} at the cost of retraining or pipeline changes; sparse attention instead reduces the attention call itself, runs on released weights, and is complementary to token reduction.

\paragraph{Sparse Attention}
Block-sparse attention, developed for long-context LLM prefill~\cite{jiang2024minference, tang2024quest, zhang2025spargeattn, xu2025xattention, lai2025flexprefill} and extended to video diffusion~\cite{xi2025svg1, yang2025svg2, zhou2026svg_ear}, computes exact attention over a selected subset of key blocks, and selection has relied on fixed structural patterns~\cite{jiang2024minference, xi2025svg1}, attention probability mass~\cite{tang2024quest, zhang2025spargeattn, xu2025xattention, lai2025flexprefill}, or the dispersion of attention logits within a block~\cite{zhou2026svg_ear}. For 3D ViTs, SparseVGGT~\cite{wang2026sparseVGGT} keeps blocks with large mass on a pooled block-level attention map via top-$k$ and top-$p$ criteria, and HeSS~\cite{kim2026hess} redistributes block budgets across heads. Whatever the domain, these criteria are statistics of the attention call itself.

\paragraph{Output-Aware KV Cache Eviction}
To our knowledge, the closest attempts to score removal by its effect on the output come from KV cache eviction, where recent methods rank each token by the change its eviction induces in the attention output~\cite{goel2026caote, guo2024vatp, feng2026criticalkv, gu2025obcache}. These scores remain per-token and stop at the attention output; ReSS minimizes the drift of the drop set as a whole, measured in the residual stream.
\begin{figure}[t]
    \centering
    \includegraphics[width=\linewidth]{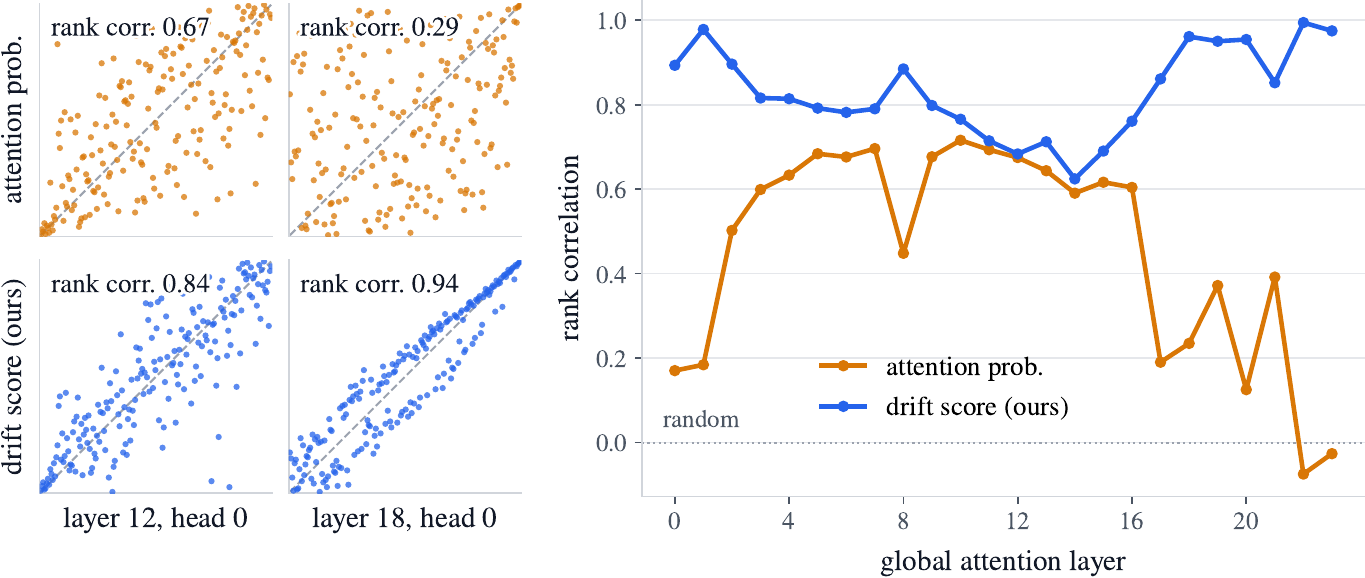}
    \caption{
    \textbf{Rank agreement with masking-induced residual changes.}
    (Left) Rank of each selection score (vertical) against the rank of the actual residual
    change caused by masking each block (horizontal); the diagonal indicates perfect
    agreement. (Right) Rank correlation across global attention layers.
    }
    \label{fig:motivation}
\end{figure}
\begin{figure*}[htbp]
    \centering
    \includegraphics[width=\linewidth]{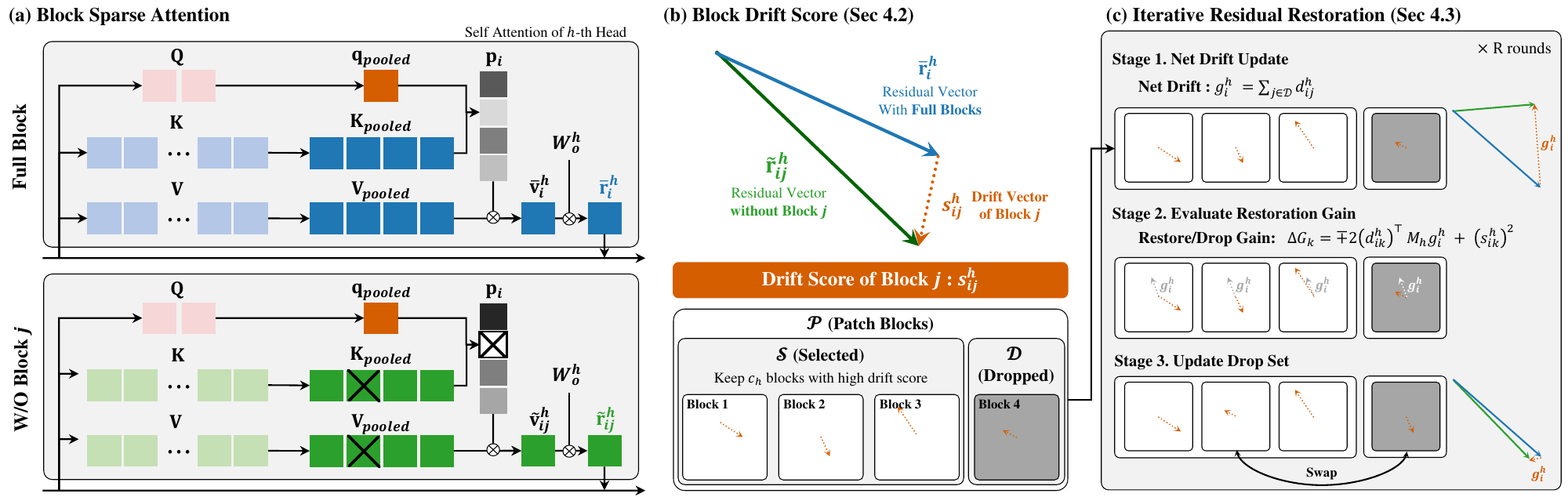}
    \caption{\textbf{Overview of ReSS.}
(a) Masking a key block shifts what attention writes into the residual stream.
(b) The drift score measures this shift per block; the $c_h$ blocks with the
highest scores form the initial selected set.
(c) Iterative restoration then reduces the drift of the drop set as a whole,
swapping blocks while preserving the budget.}
    \label{fig:overall_framework}
\end{figure*}

\section{Motivation}
\label{sec:motivation}
\paragraph{Block-Sparse Attention for 3D ViTs}
For 3D ViTs, SparseVGGT~\cite{wang2026sparseVGGT} and HeSS~\cite{kim2026hess} adopt block-sparse attention, which proceeds in three steps: (1) queries and keys are partitioned into fixed-size blocks, and mean pooling within each block yields a coarse block-level attention map; (2) for each query block and head, key blocks are selected based on this map, up to a per-head budget $c_h$; (3) exact attention is computed only over the selected blocks, with softmax renormalized among them, so the output behaves as if the unselected blocks never existed. Special tokens such as camera and register tokens aggregate global information and are excluded from sparsification~\cite{wang2026sparseVGGT}. The block layout of step (1) also affects the approximation~\cite{yang2025svg2}, but given a layout, its quality is governed by step (2): which blocks to keep. This selection criterion is the subject of this paper.

\paragraph{The Probability Assumption}
Existing methods, including SparseVGGT and HeSS, uniformly select blocks by attention probability. This choice carries an implicit assumption: the less attention a block receives, the less the model's output changes when it is removed. The assumption is typically taken for granted, and the criterion has rarely been explored beyond probability.

\paragraph{Testing the Probability Assumption}
We put this assumption to the test by directly measuring the change that masking a single block induces. For each candidate block, we run two forward passes, one with dense attention and one with only that block masked, and take the difference between the resulting attention outputs after the output projection. The magnitude of this difference, a Euclidean norm over the query tokens of the block, is our reference: it is the change that masking actually leaves in the residual stream, measured at token resolution, and its definition involves no selection criterion. We then compare the ranking assigned by attention probability against the ranking induced by this reference, computing the Spearman rank correlation for every (query block, head) pair on six DTU~\cite{aanaes2016dtu} scenes with twelve views each.

\Cref{fig:motivation} shows that attention probability fails to consistently predict this ranking. At layer 18, head 0, its rank correlation is only $0.29$; across layers, the correlation fluctuates widely and approaches or even falls below random selection in later layers. Attention probability is thus an incomplete proxy for block importance.

\paragraph{The Right Objective}
Since probability fails to predict the change, we take the change itself as the objective. Block masking changes only what the attention sublayer adds to the residual stream; the entire cost of sparsification is a shift in this contribution. We call this shift the \emph{residual drift}. Writing the pre-LN structure as $x_{\ell+1} = x_\ell + f_\ell(x_\ell)$, with $x_\ell$ the input to layer $\ell$ and $f_\ell$ its sublayer, the drift injected at layer $\ell$ propagates as a change in the input to every subsequent layer. To keep the output of the sparse model close to that of the dense model, block selection should therefore minimize the residual drift. This is the selection objective of this paper.
\section{Method}
\label{sec:method}

We propose \emph{ReSS} (\textbf{Res}idual-\textbf{Res}toring \textbf{S}parse Attention), a block selection method that minimizes residual drift. The overall pipeline is shown in \cref{fig:overall_framework}.

\subsection{Problem Formulation}
\label{sec:problem-formulation}

$P_{ij}^h$ denotes the pooled attention probability from query block $i$ to key block $j$ in head $h$, and $V_j^h$ the corresponding pooled value. Let $\mathcal{B}$ be the set of all key blocks, over which the pooled attention map is normalized, so that the dense pooled attention output is

\begin{equation}
\bar{v}_i^h
=
\sum_{j \in \mathcal{B}} P_{ij}^h V_j^h.
\label{eq:dense-pooled}
\end{equation}
Blocks containing special tokens are always kept (\cref{sec:motivation}); the remaining blocks, consisting only of patch keys, form the candidate set $\mathcal{P} \subseteq \mathcal{B}$.

For each query block $i$ and head $h$, we formulate block selection as choosing a drop set $\mathcal{D}_i^h \subseteq \mathcal{P}$, whose complement $\mathcal{S}_i^h = \mathcal{P} \setminus \mathcal{D}_i^h$ is the selected set:
\begin{equation}
    \min_{\mathcal{D}_i^h \subseteq \mathcal{P},\;
          \lvert \mathcal{D}_i^h \rvert = \lvert \mathcal{P} \rvert - c_h}
    \; J_i^h(\mathcal{D}_i^h),
    \label{eq:problem}
\end{equation}
where $J_i^h$ is the squared magnitude of the drift that the attention sublayer leaves in the residual stream when $\mathcal{D}_i^h$ is removed, and $c_h$ is the per-head budget; the constraint gives $\lvert \mathcal{S}_i^h \rvert = c_h$. The problem is solved independently for each (query block, head) pair.

\Cref{eq:problem} is a combinatorial optimization over subsets of $\mathcal{P}$, whose search space grows exponentially with the number of candidates; solving it exactly within a forward pass at every layer is infeasible. ReSS approximates it in two stages: \cref{sec:block-drift-score} constructs an initial selected set from a per-block drift score (\cref{fig:overall_framework}b), and \cref{sec:iterative-restoration} directly lowers $J_i^h$ through iterative restoration (\cref{fig:overall_framework}c).

\subsection{Block Drift Score}
\label{sec:block-drift-score}

\paragraph{Drift induced by block masking.}
We first consider masking a single key block $j$. Since softmax is renormalized over the remaining blocks, the attention output after masking is
\begin{equation}
    \tilde{v}_{ij}^h
    =
    \sum_{k \in \mathcal{B},\, k \ne j}
    \frac{P_{ik}^h}{1 - P_{ij}^h} V_k^h
    =
    \frac{\bar{v}_i^h - P_{ij}^h V_j^h}
    {1 - P_{ij}^h},
    \label{eq:masked-pooled-attention}
\end{equation}
and the resulting output change is exactly
\begin{equation}
    \tilde{v}_{ij}^h - \bar{v}_i^h
    =
    \frac{P_{ij}^h}{1 - P_{ij}^h}
    \left( \bar{v}_i^h - V_j^h \right).
    \label{eq:single-block-drift}
\end{equation}
This change can be decomposed into three factors: the deviation $e_{ij}^h = \bar{v}_i^h - V_j^h$ of the block's value from the dense output, the attention probability $P_{ij}^h$, and the factor $1/(1-P_{ij}^h)$ from softmax renormalization. For the subsequent development, we set the renormalization factor aside and define the unnormalized drift contribution
\begin{equation}
    d_{ij}^h = P_{ij}^h e_{ij}^h.
    \label{eq:unnormalized-drift}
\end{equation}

\paragraph{Mapping to the residual space.} The attention output enters the residual stream through the output projection, so the impact of masking must be measured after $W_O^h \in \mathbb{R}^{d \times d_h}$, the slice of the output projection for head $h$, with residual dimension $d$ and head dimension $d_h$. We define the unnormalized block drift score
\begin{equation}
s_{ij}^h = \left\lVert W_O^h d_{ij}^h \right\rVert_2 = P_{ij}^h \left\lVert W_O^h e_{ij}^h \right\rVert_2.
\label{eq:block-drift-score}
\end{equation}
For selection, we restore the renormalization factor and keep the $c_h$ blocks with the largest exact single-block drift magnitude,
\begin{equation}
\frac{s_{ij}^h}{\max(1-P_{ij}^h, \epsilon)}, \qquad \epsilon = 10^{-4}.
\label{eq:cut-key}
\end{equation}

\paragraph{Efficient score computation.} 
Computing \cref{eq:block-drift-score} directly materializes $W_O^h e_{ij}^h$ for every head and query--key block pair, a prohibitive intermediate of size $(n_h, q_{\mathrm{blk}}, k_{\mathrm{blk}}, d)$ over $n_h$ heads, $q_{\mathrm{blk}}$ query blocks, and $k_{\mathrm{blk}}$ key blocks. 
We instead precompute the per-head Gram matrix
\begin{equation}
    M_h = (W_O^h)^\top W_O^h
    \in \mathbb{R}^{d_h \times d_h}
    \label{eq:gram}
\end{equation}
once at model load time and expand the squared projected deviation as
\begin{equation}
\begin{aligned}
    q_{ij}^h
    &= \left\lVert W_O^h e_{ij}^h \right\rVert_2^2
    = (e_{ij}^h)^\top M_h e_{ij}^h \\
    &= (\bar{v}_i^h)^\top M_h \bar{v}_i^h
    - 2(\bar{v}_i^h)^\top M_h V_j^h
    + {V_j^h}^\top M_h V_j^h,
\end{aligned}
\label{eq:projected-deviation}
\end{equation}
three low-dimensional contractions with no per-pair vector intermediate; $s_{ij}^h = P_{ij}^h \sqrt{q_{ij}^h}$ then recovers \cref{eq:block-drift-score} exactly.

\subsection{Iterative Residual Restoration}
\label{sec:iterative-restoration}

The initial selection of \cref{eq:cut-key} evaluates each block independently. However, the drift of removing multiple blocks is determined by the sum of their contribution vectors, in which opposing directions partially offset. ReSS therefore refines the initial selection iteratively, lowering the drift of the drop set as a whole (\cref{fig:overall_framework}c).

\begin{figure*}[htbp]
    \centering
    \includegraphics[width=\linewidth]{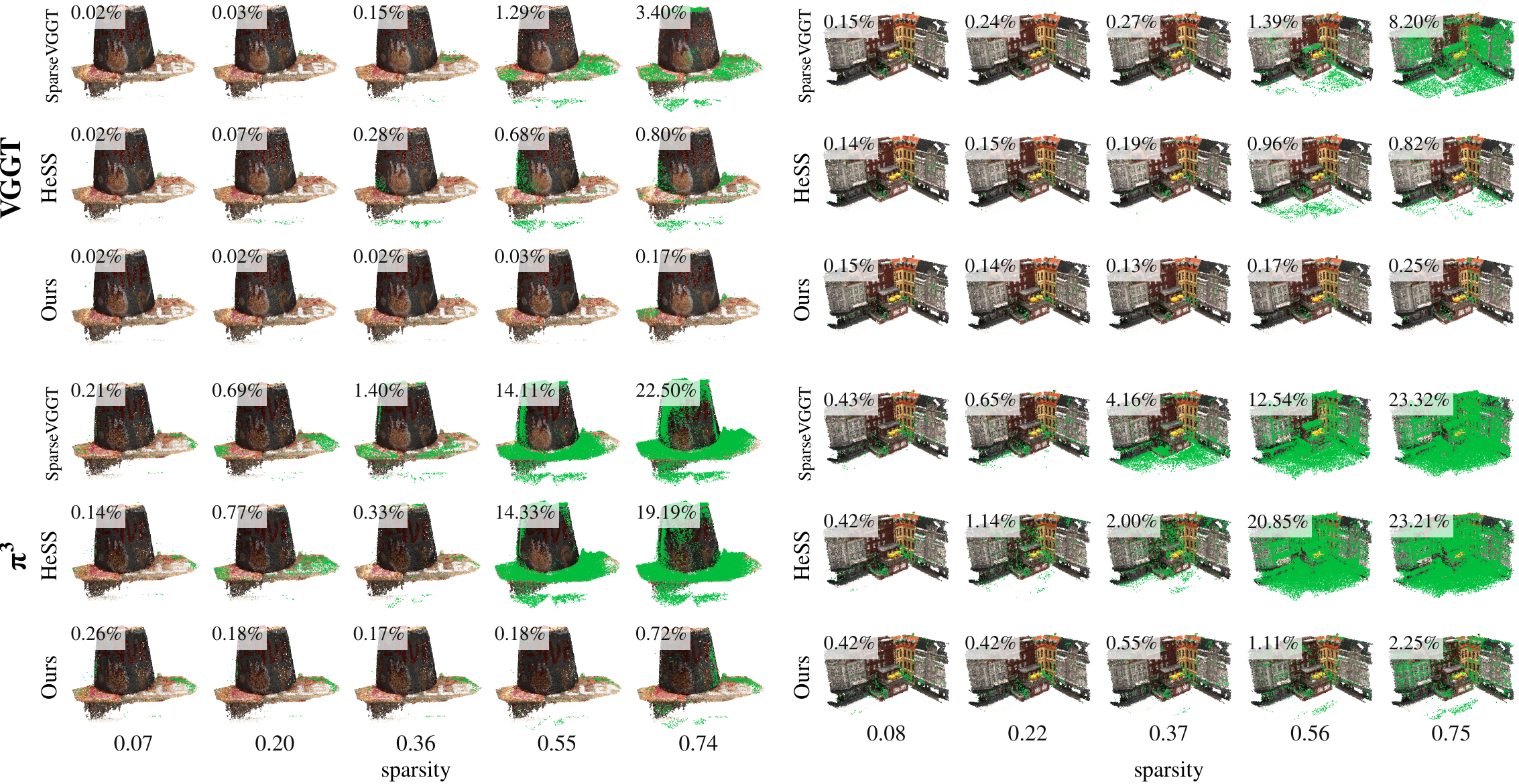}
    \vspace{-2em}
    \caption{\textbf{Qualitative results.}
    Two DTU scans reconstructed at five sparsity levels, on VGGT~\cite{wang2025vggt} and $\mathbf{\pi^3}$~\cite{wang2026pi}.
    Columns are the measured sparsity, matched across methods. Points are drawn in green where their distance to the ground truth exceeds 5\,mm, and each panel reports the fraction of such points.
    }
    \label{fig:qual_res}
\end{figure*}

\paragraph{Net drift of the drop set.}
Let $\mathcal{D}$ be the current drop set. We define the net drift and the remaining attention probability mass as
\begin{equation}
    g_i^h = \sum_{j \in \mathcal{D}} d_{ij}^h,
    \qquad
    K_i^h = 1 - \sum_{j \in \mathcal{D}} P_{ij}^h.
    \label{eq:net-drift}
\end{equation}
Since $\mathcal{D} \subseteq \mathcal{P}$ and the blocks in $\mathcal{B} \setminus \mathcal{P}$ are always kept, $K_i^h$ stays away from zero. The attention output after masking all blocks in $\mathcal{D}$ is
\begin{equation}
    \tilde{v}_i^h
    = \bar{v}_i^h + \frac{g_i^h}{K_i^h},
    \label{eq:masked-output}
\end{equation}
and the resulting squared residual drift is
\begin{equation}
    J_i^h(\mathcal{D})
    = \left\lVert W_O^h \frac{g_i^h}{K_i^h} \right\rVert_2^2
    = \frac{(g_i^h)^\top M_h g_i^h}{(K_i^h)^2}.
    \label{eq:net-drift-objective}
\end{equation}
This is the objective of \cref{eq:problem}. We denote its numerator by $G_i^h = (g_i^h)^\top M_h g_i^h$, the squared magnitude of the net contribution measured in the residual space. For $\mathcal{D}=\{j\}$, \cref{eq:net-drift-objective} reduces to the square of the ranking key in \cref{eq:cut-key}; the initial selection and the iterative refinement thus minimize the same objective at different set sizes, rather than two different ones.

\paragraph{Budget-neutral swaps.}
Starting from the initial drop set, ReSS repeats budget-neutral swaps: each round first restores the blocks that most reduce the current net drift, then drops an equal number of blocks to preserve the budget.

Restoring a block $k \in \mathcal{D}$ updates
\begin{equation}
    g_i^h \leftarrow g_i^h - d_{ik}^h,
    \qquad
    K_i^h \leftarrow K_i^h + P_{ik}^h,
    \label{eq:restore-update}
\end{equation}
and changes the numerator of \cref{eq:net-drift-objective} by
\begin{equation}
    \Delta G_k^{+}
    = -2(d_{ik}^h)^\top M_h g_i^h + (s_{ik}^h)^2.
    \label{eq:restore-delta}
\end{equation}
Conversely, dropping a kept block $k \notin \mathcal{D}$ changes it by
\begin{equation}
    \Delta G_k^{-}
    = 2(d_{ik}^h)^\top M_h g_i^h + (s_{ik}^h)^2.
    \label{eq:drop-delta}
\end{equation}
Denoting the change in $G_i^h$ from one update by $\Delta G$ and that in $K_i^h$ by $u$ ($u = +P_{ik}^h$ for a restore, $-P_{ik}^h$ for a drop), the exact decrease of the objective is
\begin{equation}
    \Delta J
    =
    \frac{G_i^h}{(K_i^h)^2}
    -
    \frac{G_i^h+\Delta G}{(K_i^h+u)^2}.
    \label{eq:objective-delta}
\end{equation}
In each round, we select the blocks to restore and, in compensation, the blocks to drop using \cref{eq:objective-delta}.

The swap budget is halved every round. Setting the total budget to a fraction $\rho$ of the keep budget, round $r$ performs
\begin{equation}
    m_r = \rho \, 2^{-(r+1)} c_h,
    \qquad r = 0, \dots, R-1,
    \label{eq:swap-schedule}
\end{equation}
swaps, for a total of $\rho(1 - 2^{-R})c_h$ over $R$ rounds. We use $\rho = 0.5$ and $R = 5$, giving $0.484\,c_h$ swaps in total. The budget is proportional to the keep budget $c_h$ rather than the drop set size: once sparsity exceeds $2/3$, $\lvert \mathcal{D} \rvert > 2c_h$, and a budget proportional to $\lvert \mathcal{D} \rvert$ could replace the entire kept set.
Throughout the iteration, ReSS tracks the drop set with the smallest objective \cref{eq:net-drift-objective} observed so far and returns it as the final selection. The objective after refinement therefore never exceeds that of the initial selection.

\paragraph{Fused kernel implementation.}
Each restoration round launches many small operations over $(n_h, q_{\mathrm{blk}}, k_{\mathrm{blk}})$ tensors; each is cheap in FLOPs but incurs a separate round trip to memory, so the round is bound by memory bandwidth rather than computation. We therefore fuse the round body into a single Triton~\cite{tillet2019triton} kernel: each query row is processed in registers in one sweep, with no intermediate tensor written to memory and the same selection as the unfused path, bit for bit. Its effect on the selection cost is reported in \cref{tab:cost_analysis}, and implementation details are in \cref{sec:supp_fused_kernel_impl}.

\subsection{Pipeline Configuration}
\label{sec:pipeline-configuration}

\paragraph{Sparsity control.}
We control sparsity with the top-$k$ budget alone. Unlike a CDF threshold, whose keep count varies with the data, a fixed budget makes the realized sparsity match its target; removing the threshold has no consistent effect on accuracy (\cref{sec:ablation}), and all experiments are compared at matched measured sparsity.

\paragraph{Head-wise budget allocation.}
ReSS assumes the per-head budgets $\{c_h\}$ as given and is compatible with any allocation scheme. Unless noted otherwise, we adopt the head-wise allocation of HeSS~\cite{kim2026hess} (see \cref{sec:supp_hess_impl}); to isolate its effect, we also evaluate a ReSS variant with uniform budgets.
\section{Experiments}
\label{sec:experiments}

\begin{figure*}[htbp]
    \centering
    \includegraphics[width=\linewidth]{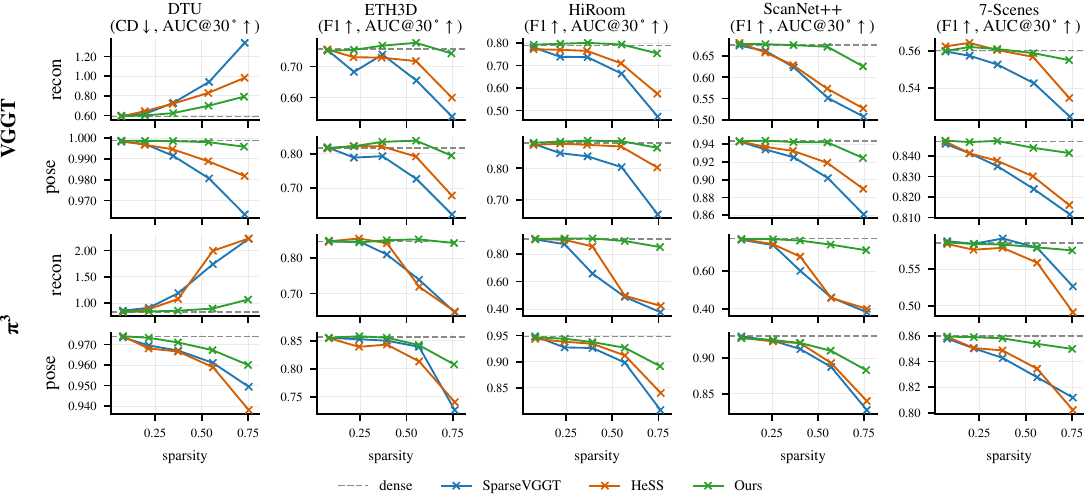}
    \vspace{-2em}
    \caption{\textbf{Quantitative results.} Reconstruction (recon) and camera pose estimation (pose) across sparsity levels, on the VGGT~\cite{wang2025vggt} (top) and $\pi^3$~\cite{wang2026pi} (bottom) backbones. Dashed lines denote dense attention; arrows mark whether higher ($\uparrow$) or lower ($\downarrow$) is better.
    As sparsity grows, \textcolor{mygreen}{ReSS} preserves dense performance better than \textcolor{myblue}{SparseVGGT}~\cite{wang2026sparseVGGT} and \textcolor{myorange}{HeSS}~\cite{kim2026hess}.}
    \label{fig:quan_res}
\end{figure*}

\subsection{Experimental Setup}
\label{sec:exp-setup}
\paragraph{Backbones.}
We evaluate on three feed-forward 3D ViTs: VGGT~\cite{wang2025vggt}, $\pi^3$~\cite{wang2026pi}, and DepthAnything3~\cite{lin2025depthanything3}. All three alternate per-frame attention with global attention over the concatenated tokens of every view, and ReSS sparsifies only the global layers: $24$ of $48$ in VGGT, $18$ of $36$ in $\pi^3$, and $14$ of $40$ in DepthAnything3. None is retrained; ReSS runs on the released weights. The main text reports VGGT and $\pi^3$; DepthAnything3 results are in \cref{sec:supp_addtional_exp}.

\paragraph{Baselines.}
Our main baselines are SparseVGGT~\cite{wang2026sparseVGGT} and HeSS~\cite{kim2026hess}, the block-sparse attention methods proposed for 3D ViTs. We additionally port six sparse attention methods from other domains to VGGT: XAttention~\cite{xu2025xattention}, SpargeAttn~\cite{zhang2025spargeattn}, and FlexPrefill~\cite{lai2025flexprefill} from LLM long-context prefill, and SVG~\cite{xi2025svg1}, SVG2~\cite{yang2025svg2}, and SVG-EAR~\cite{zhou2026svg_ear} from video diffusion. The LLM methods share our block configuration and kernel, differing only in the selection criterion; the video methods keep their own block layouts (\cref{sec:supp_llm_baselines}, \cref{sec:supp_svg_baselines}).

\paragraph{Benchmarks.}
We measure 3D geometry performance on DA3-Bench, the evaluation pipeline of DepthAnything3~\cite{lin2025depthanything3}, which covers HiRoom~\cite{lin2025depthanything3}, 7-Scenes~\cite{shotton20137scenes}, DTU~\cite{aanaes2016dtu}, ETH3D~\cite{schops2017eth3d}, and ScanNet++~\cite{yeshwanth2023scannet++}, evaluating camera pose estimation and multi-view geometry on each. Each scene is evaluated on all of its frames, subsampled to at most $100$: HiRoom has $10$--$23$ views, ETH3D $14$--$76$, DTU $49$, and ScanNet++ and 7-Scenes reach the cap. To measure the degradation precisely across sparsity levels, we add an alignment step based on dense point correspondences to the evaluation pipeline (\cref{sec:supp_dense_alignment}). 

\paragraph{Metrics.}
Multi-view geometry is reported as Chamfer distance ($\downarrow$) on DTU and F1 score ($\uparrow$) elsewhere, with distance threshold $\tau = 0.25\,\mathrm{m}$ on ETH3D and $0.05\,\mathrm{m}$ on the other datasets. Camera pose is reported as AUC@$30^\circ$ ($\uparrow$) over relative rotation and translation errors. Exact definitions follow DA3-Bench~\cite{lin2025depthanything3}.

\subsection{Qualitative Results}
\label{sec:qual_res}
\Cref{fig:qual_res} shows reconstructions of two DTU scans at five sparsity levels, with points more than $5$\,mm from the ground truth drawn in green. As sparsity grows, the baselines progressively lose the scene structure, while ReSS keeps it largely intact at every level. The contrast is sharpest at the highest sparsity on $\pi^3$: on the right scan, errors spread across the scene for SparseVGGT and HeSS, with $23.32\%$ and $23.21\%$ of points over the threshold, while ReSS stays at $2.25\%$. The reconstructions thus make visible the gap that \cref{sec:quan_res} quantifies.

\subsection{Quantitative Results}
\label{sec:quan_res}
\Cref{fig:quan_res} shows performance across sparsity levels; dashed lines mark dense performance. All methods match dense at low sparsity, but as compression grows, SparseVGGT and HeSS depart from the dashed lines while ReSS stays on them up to sparsity $0.4$. The gap is widest at the highest operating point: on $\pi^3$/DTU, the baselines' Chamfer distance exceeds twice the dense value while ReSS remains near dense ($1.067$ vs.\ $0.838$), and on $\pi^3$/HiRoom, F1 falls to $0.378$ and $0.424$ while ReSS keeps $0.850$ (dense $0.908$). Pose degrades more gently but in the same order. At the highest operating point, ReSS is best in both reconstruction and pose in all ten backbone--dataset combinations, and a similar trend holds on DepthAnything3 (\cref{sec:supp_da3}).

\begin{figure}[t]
    \centering
    \includegraphics[width=\linewidth]{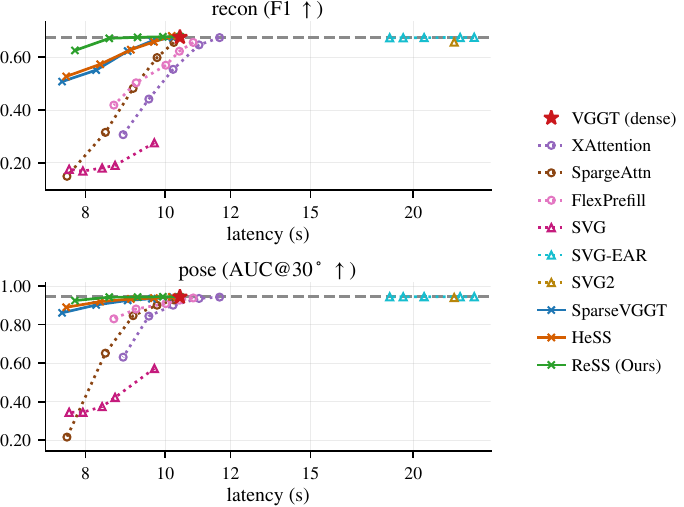}
    \vspace{-1.6em}
    \caption{\textbf{Quality against latency.} Each point is one operating point on ScanNet++ with the VGGT backbone; the dashed line marks dense attention. Latency covers the full forward pass including all selection overhead. LLM criteria reach dense quality only at dense latency or beyond; video methods lose either quality (SVG) or speed (SVG2, SVG-EAR).}
    \label{fig:latency_quality}
\end{figure}

\subsection{Sparse Attention from Other Domains}
\cref{fig:latency_quality} compares the six ported methods against ReSS in the latency--quality plane.
The three LLM criteria trace the same shape: all collapse in the compression regime, where speed gains actually appear. For XAttention and FlexPrefill the masks show why: each sizes its mask by a threshold on attention mass alone, so at matched sparsity the loss concentrates on a few rows rather than spreading evenly (\cref{sec:supp_llm_baselines}).

Video diffusion appears the closer neighbor, as both consume long token sequences from many images, yet the transfer fails on two counts. First, the assumptions do not carry: the fixed bands of SVG assume adjacent frames share nearly the same scene, but adjacent views are much farther apart even on ScanNet++, where the view order follows a capture trajectory, and SVG reaches dense quality at no operating point. Second, the costs do not amortize: the clustering of SVG2 and SVG-EAR carries over, and SVG-EAR retains dense quality even at high sparsity, yet both run slower than dense attention, since re-planning the block layout at every layer is a fixed cost that diffusion spreads over tens of denoising steps and a single forward pass pays in full (\cref{sec:supp_svg_baselines}).

\begin{figure}[t]
    \centering
    \includegraphics[width=\linewidth]{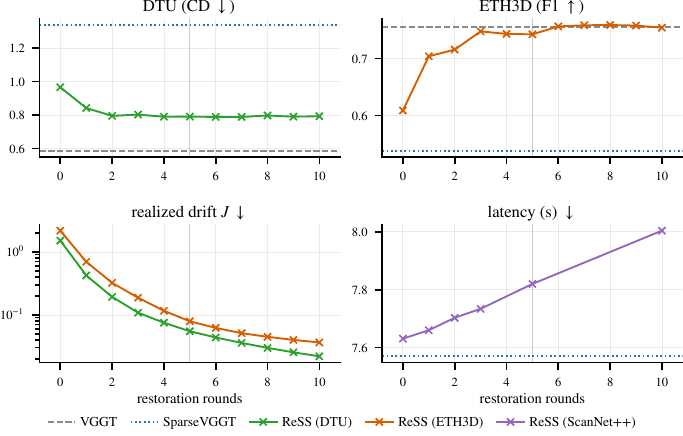}
    \vspace{-1.6em}
    \caption{\textbf{Number of restoration rounds.} Quality saturates after three to four rounds, while the realized drift keeps decreasing beyond that point. The grey line marks the setting we adopt. Dashed line denotes dense attention.}
    \label{fig:niters}
\end{figure}

\begin{figure}[t]
    \centering
    \includegraphics[width=\linewidth]{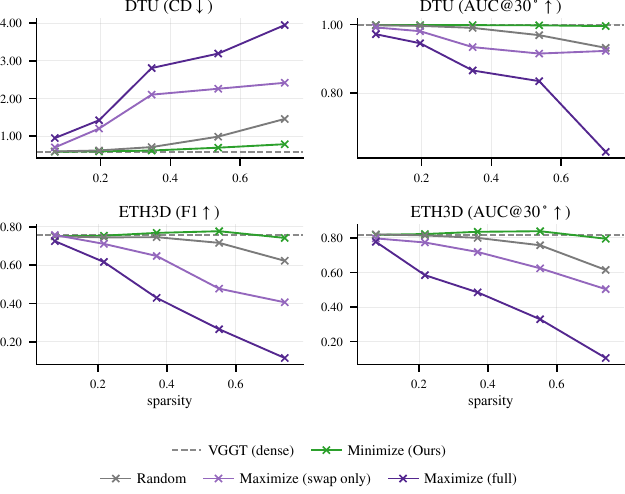}
    \vspace{-1.6em}
    \caption{\textbf{Drift minimization vs maximization.} With the budget and sparsity fixed, we flip the sign of the objective, either
only in the restoration stage (swap only) or from the initial selection (full).
Maximizing drift collapses performance faster than random selection,
supporting drift as a valid selection criterion.}
    \label{fig:drift_min_max}
\end{figure}

\section{Analysis}
\subsection{Does Drift Explain the Cost of Sparsification?}
\label{sec:drift-analysis}
We examine the premise of ReSS from three directions: whether ReSS actually reduces drift, whether reducing drift preserves quality, and what reducing it costs.

\paragraph{Does ReSS reduce drift?}
\Cref{fig:niters} varies the number of restoration rounds from $0$ to $10$ with all else fixed. The realized drift $J$ falls by more than an order of magnitude from the initial selection, confirming that the iterative refinement optimizes what it claims to.

\begin{figure}[t]
    \centering
    \includegraphics[width=\linewidth]{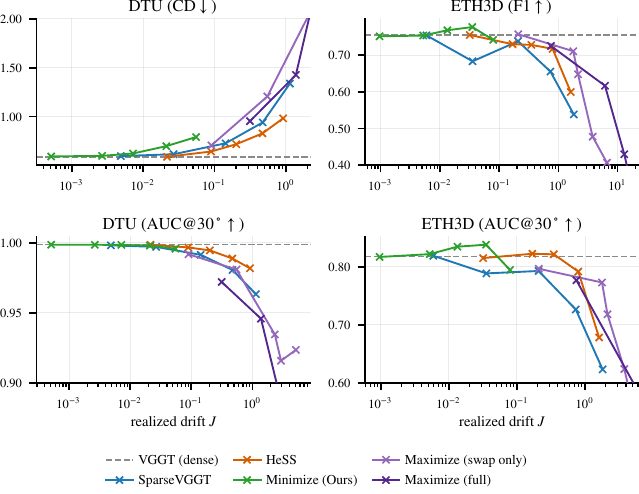}
    \vspace{-1.6em}
    \caption{\textbf{Drift determines quality.} Performance plotted against the realized drift $J$ of each selection instead of sparsity, on the VGGT~\cite{wang2025vggt} backbone over DTU and ETH3D. Different methods, including the drift-maximizing variants, collapse onto a single curve.
}
    \label{fig:drift_quality}
\end{figure}

\paragraph{Does reducing drift preserve quality?}
\Cref{fig:drift_min_max} flips the sign of the objective with the budget and sparsity fixed: maximizing drift collapses performance even at mild sparsity, far below random selection. \Cref{fig:drift_quality} plots quality against the realized drift of each selection, and all methods land on nearly the same curve, the drift-maximizing variants included. Methods that differ widely at equal sparsity nearly coincide at equal drift.

Together, the two results say more than that minimizing drift works: performance tracks the drift a selection leaves, regardless of how the selection is made.

\begin{table}[t]
    \centering
    \small
    \setlength{\tabcolsep}{3pt}
    \caption{\textbf{Cost analysis.} Selection cost at $100$ views on ScanNet++, sparsity $0.745$. Parenthesized: overhead over SparseVGGT~\cite{wang2026sparseVGGT}. RTX 4090, bfloat16.}
    \label{tab:cost_analysis}

    \resizebox{\linewidth}{!}{
    \begin{tabular}{lccc}
    \toprule
     & Latency & Scoring mem. & Peak mem. \\
                  & (s)     & (MiB)        & (GiB)     \\
    \midrule
    SparseVGGT               & $7.58$          & ---    & --- \\
    \midrule
    ReSS (full)              & $\textbf{7.82}$ ($\textbf{+0.24}$) & $\textbf{205}$  & $\textbf{16.4}$ \\
    w/o fused kernel   & $8.84$ ($+1.26$) & $\textbf{205}$  & $\textbf{16.4}$ \\
    w/o expansion      & $8.48$ ($+0.90$) & $7760$ & $22.4$ \\
    w/o cached $M_h$   & $7.87$ ($+0.29$) & $278$  & $\textbf{16.4}$ \\
    \bottomrule
    \end{tabular}
    }
\end{table}

\paragraph{What does reducing drift cost?}
With $R=5$ rounds, the entire selection path, scoring and restoration included, adds $0.24$\,s over SparseVGGT and $205$\,MiB of scoring memory at sparsity 0.745. Five rounds suffice: quality saturates after three to four rounds while each additional round adds a near-constant latency (\cref{fig:niters}). The cost stays this low because of two reformulations. Expanding the quadratic form in \cref{eq:projected-deviation} removes the per-pair $d_h$-dimensional intermediate, which otherwise grows the scoring allocation $38\times$
and raises peak memory from $16.4$ to $22.4$\,GiB. 
Keeping each round in registers with a fused kernel accounts for the latency: without it, the overhead grows about $5 \times$, to $+1.26$\,s. \Cref{tab:cost_analysis} also reports the smaller saving from caching $M_h$.

\begin{table}[t]
    \centering
    \small
    \caption{\textbf{Ablation study.} Components are removed ($-$) or added ($+$) cumulatively, one per row, from ReSS down to SparseVGGT~\cite{wang2026sparseVGGT}; the last row coincides with SparseVGGT. VGGT~\cite{wang2025vggt} backbone, all rows matched at the highest operating point.}
    \setlength{\tabcolsep}{4pt}
    \resizebox{\linewidth}{!}{
    \begin{tabular}{lcccc}
        \toprule
        & \multicolumn{2}{c}{Reconstruction} & \multicolumn{2}{c}{Pose (AUC@$30^\circ$)} \\
        \cmidrule(lr){2-3} \cmidrule(lr){4-5}
        & DTU $\downarrow$ & ETH3D $\uparrow$ & DTU $\uparrow$ & ETH3D $\uparrow$ \\
        \midrule
        ReSS (full)                                    & \textbf{0.791} & \textbf{0.743} & \textbf{0.996} & \textbf{0.795} \\
        \quad $-$ head budget                          & 0.983 & 0.696 & 0.980 & 0.762 \\
        \quad $-$ restoration                          & 1.151 & 0.582 & 0.980 & 0.702 \\
        \quad $-$ drift score                          & 1.481 & 0.581 & 0.958 & 0.643 \\
        \quad $+$ CDF threshold                        & 1.338 & 0.538 & 0.963 & 0.623 \\

        \bottomrule
    \end{tabular}
    }
    \label{tab:ablation}
\end{table}
\subsection{Ablation Study}
\label{sec:ablation}
\Cref{tab:ablation} strips ReSS down to SparseVGGT one component at a time, cumulatively, so that the last row coincides with SparseVGGT; all rows are matched at a measured sparsity of $0.73$ on the VGGT backbone.

Each row tests one claim. Taking away head-wise budget allocation costs a consistent margin on all four metrics, showing that the criterion benefits from, but does not depend on, how the budget is distributed across heads. Removing restoration costs the largest margin on ETH3D, in both reconstruction and pose, which follows from its premise: the drift of a drop set is the sum of contribution vectors, so independent scoring leaves cancellation that only set-level refinement recovers (\cref{sec:iterative-restoration}). Falling back from the drift score to probability selection costs the largest margin on DTU, in both reconstruction and pose, without changing how many blocks are kept, only which (\cref{sec:block-drift-score}). Restoring the CDF threshold on top changes results in neither direction consistently, confirming it can be dropped for simpler sparsity control without paying in quality (\cref{sec:pipeline-configuration}).

Every component contributes, including the two that embody our claim: drift scoring and set-level restoration.

\section{Conclusion}
\label{sec:conclusion}
We recast block selection in 3D vision transformers, from keeping blocks with high attention probability to minimizing the drift that sparsification leaves in the residual stream. ReSS realizes this with a drift score for single-block removal and iterative residual restoration that reduces the cumulative drift of the drop set. Across three backbones and five datasets, ReSS preserves dense performance better than prior methods at matched sparsity; flipping the sign of the objective shows that this gain comes from the drift criterion itself, and performance across methods collapses onto a single curve as a function of the drift each leaves. Sparsity says how much attention is skipped; the drift left in the residual stream says what that skipping costs, and selection should be designed, and perhaps evaluated, in that currency.

\section*{Acknowledgements}
This work was supported by Samsung Electronics Co., Ltd [No. IO260120-15267-01];
Information \& communications Technology Planning \& Evaluation (IITP) grant funded by the Korea government (MSIT) [No. RS-2021-II211343; RS-2022-II220959; RS-2025-02263754; Artificial Intelligence Graduate School Program (Seoul National University)]; 
the National Research Foundation of Korea (NRF) grant funded by MSIT [No. 2022R1A3B1077720; 2022R1A5A7083908] and the BK21 Four program of the Education and Research Program for Future ICT Pioneers, SNU in 2026.
This research was also conducted as part of the Sovereign AI Foundation Model Project (Data Track), organized by MSIT and supported by the National Information Society Agency (NIA) of Korea [No. 2025-AI Data-wi43].
{
    \small
    \bibliographystyle{ieeenat_fullname}
    \bibliography{main}
}

\clearpage
\setcounter{page}{1}
\setcounter{section}{0}
\renewcommand\thesection{\Alph{section}}
\setcounter{table}{0}
\renewcommand{\thetable}{S\arabic{table}}
\setcounter{figure}{0}
\renewcommand{\thefigure}{S\arabic{figure}}
\maketitlesupplementary

\begin{table*}[t]
\centering\small
\caption{\textbf{Absolute numbers behind \cref{fig:quan_res} of the main paper.} Three of
the five operating points, showing the reconstruction metric (Chamfer on DTU, F1
elsewhere). The best value within each compression group is in bold. Sparsity is
measured, not requested.}
\label{tab:supp_quan_res}
\resizebox{\textwidth}{!}{\begin{tabular}{llrrrrrrrrrr}
\toprule
& & dense & \multicolumn{3}{c}{$s\!=\!0.07$} & \multicolumn{3}{c}{$s\!=\!0.35$} & \multicolumn{3}{c}{$s\!=\!0.73$} \\
& & & SparseVGGT & HeSS & Ours & SparseVGGT & HeSS & Ours & SparseVGGT & HeSS & Ours \\
\midrule
VGGT & DTU $\downarrow$ & 0.588 & 0.597 & \textbf{0.590} & 0.593 & 0.727 & 0.718 & \textbf{0.623} & 1.338 & 0.982 & \textbf{0.791} \\
VGGT & ETH3D $\uparrow$ & 0.756 & 0.754 & \textbf{0.756} & 0.751 & 0.739 & 0.729 & \textbf{0.768} & 0.538 & 0.600 & \textbf{0.743} \\
VGGT & HiRoom $\uparrow$ & 0.790 & 0.781 & 0.773 & \textbf{0.793} & 0.738 & 0.766 & \textbf{0.802} & 0.472 & 0.575 & \textbf{0.755} \\
VGGT & ScanNet++ $\uparrow$ & 0.676 & 0.674 & \textbf{0.681} & 0.678 & 0.623 & 0.629 & \textbf{0.676} & 0.508 & 0.527 & \textbf{0.626} \\
VGGT & 7-Scenes $\uparrow$ & 0.560 & 0.560 & \textbf{0.563} & 0.560 & 0.553 & 0.561 & \textbf{0.561} & 0.525 & 0.535 & \textbf{0.555} \\
$\pi^3$ & DTU $\downarrow$ & 0.838 & 0.858 & \textbf{0.827} & 0.839 & 1.192 & 1.079 & \textbf{0.857} & 2.221 & 2.227 & \textbf{1.067} \\
$\pi^3$ & ETH3D $\uparrow$ & 0.846 & 0.845 & 0.847 & \textbf{0.849} & 0.810 & 0.841 & \textbf{0.850} & 0.647 & 0.649 & \textbf{0.842} \\
$\pi^3$ & HiRoom $\uparrow$ & 0.908 & 0.905 & 0.909 & \textbf{0.909} & 0.658 & 0.855 & \textbf{0.912} & 0.378 & 0.424 & \textbf{0.850} \\
$\pi^3$ & ScanNet++ $\uparrow$ & 0.774 & 0.767 & \textbf{0.771} & 0.771 & 0.602 & 0.678 & \textbf{0.763} & 0.381 & 0.400 & \textbf{0.711} \\
$\pi^3$ & 7-Scenes $\uparrow$ & 0.586 & \textbf{0.588} & 0.585 & 0.587 & \textbf{0.592} & 0.579 & 0.584 & 0.526 & 0.491 & \textbf{0.576} \\
\bottomrule
\end{tabular}
}
\end{table*}

\begin{table*}[t]
\centering\small
\caption{\textbf{Absolute numbers for DepthAnything3.} The same runs as
\cref{fig:supp_da3}. Read as in \cref{tab:supp_quan_res}.}
\label{tab:supp_da3}
\resizebox{\textwidth}{!}{\begin{tabular}{llrrrrrrrrrr}
\toprule
& & dense & \multicolumn{3}{c}{$s\!=\!0.10$} & \multicolumn{3}{c}{$s\!=\!0.48$} & \multicolumn{3}{c}{$s\!=\!0.80$} \\
& & & SparseVGGT & HeSS & Ours & SparseVGGT & HeSS & Ours & SparseVGGT & HeSS & Ours \\
\midrule
DA3 & DTU $\downarrow$ & 0.850 & \textbf{0.852} & 0.875 & 0.854 & 0.985 & 1.464 & \textbf{0.901} & 1.625 & 2.205 & \textbf{1.195} \\
DA3 & ETH3D $\uparrow$ & 0.853 & \textbf{0.857} & 0.848 & 0.851 & 0.847 & 0.851 & \textbf{0.854} & 0.786 & \textbf{0.835} & 0.832 \\
DA3 & HiRoom $\uparrow$ & 0.894 & 0.892 & \textbf{0.893} & 0.891 & \textbf{0.877} & 0.846 & 0.855 & 0.760 & 0.718 & \textbf{0.796} \\
DA3 & ScanNet++ $\uparrow$ & 0.772 & 0.772 & \textbf{0.772} & 0.772 & 0.768 & 0.766 & \textbf{0.771} & 0.743 & 0.755 & \textbf{0.771} \\
DA3 & 7-Scenes $\uparrow$ & 0.569 & 0.570 & 0.569 & \textbf{0.570} & 0.572 & \textbf{0.572} & 0.570 & 0.564 & 0.571 & \textbf{0.572} \\
\bottomrule
\end{tabular}
}
\end{table*}

\begin{figure*}[t]
\centering
\includegraphics[width=\linewidth]{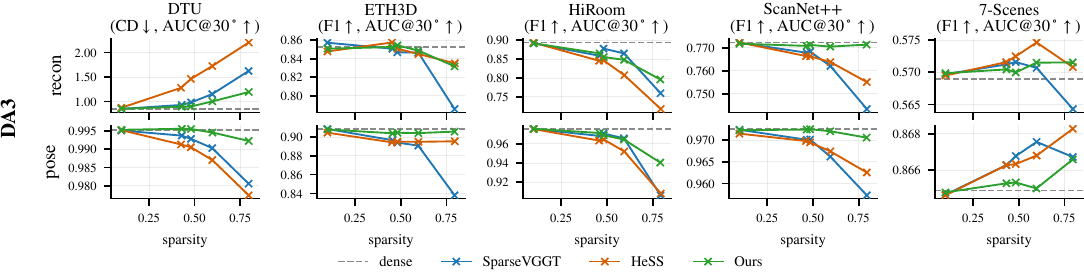}
\caption{\textbf{DepthAnything3 on the five DA3-Bench datasets.} Read as in \cref{fig:quan_res}: the vertical axis is the metric in its own units with the dashed line at dense, and the horizontal axis is measured sparsity. The trend matches the two backbones in the main paper, but the margins between methods are smaller. On 7-Scenes the methods differ only slightly, so that axis is magnified accordingly.}
\label{fig:supp_da3}
\end{figure*}

\section{Limitation}
\label{sec:supp_limitation}
The objective of ReSS is to minimize drift from the dense computation, and this takes the dense output as an attainable ceiling. The experiments show that the premise does not always hold (\cref{sec:supp_da3}). In some cases sparsification improves on dense, which means that the drift introduced at the selection step can act, as it passes through the subsequent layers of the model, in a direction that raises quality. The current formulation addresses only the magnitude of drift and does not ask whether its direction is beneficial or harmful, so ReSS stays near the dense result by design and forgoes these gains. Distinguishing beneficial from harmful drift at selection time, that is, sparsification that improves on the dense computation rather than restoring it, is worth pursuing as future work.

\section{Additional Experimental Results}
\label{sec:supp_addtional_exp}

\subsection{Additional Results for VGGT~\cite{wang2025vggt} and $\pi^3$}
\cref{tab:supp_quan_res} gives the absolute numbers behind \cref{fig:quan_res} at three of its five operating points.
\cref{fig:supp_qual1} and \cref{fig:supp_qual2} extend the qualitative comparison of
\cref{fig:qual_res} to two further DTU~\cite{aanaes2016dtu} scans and to ETH3D~\cite{schops2017eth3d}, HiRoom~\cite{lin2025depthanything3} and ScanNet++~\cite{yeshwanth2023scannet++}.

\subsection{DepthAnything3}
\label{sec:supp_da3}

\cref{fig:supp_da3} and \cref{tab:supp_da3} report the same sweep on DepthAnything3~\cite{lin2025depthanything3}: the same five datasets, the same five operating points, the same sparsity-matching procedure. Sparsification is applied to the $14$ of its $40$ blocks that perform cross-view attention.

The trend is the one in the main paper, with smaller margins between methods than on the other two backbones. At the highest operating point, ReSS gives the best reconstruction on every dataset except ETH3D, where HeSS is marginally ahead. At lower sparsity the methods lie within a small margin of one another, and where one pulls ahead, it is often by scoring above dense, that is, sparsification happens to improve on the dense computation. The one clear exception is HiRoom at intermediate sparsity, where SparseVGGT stays closer to dense than ReSS. ReSS minimizes drift away from the dense computation, so it stays near the dense result and does not gain from departures that improve on it; Sec.~\ref{sec:supp_limitation} takes up this limitation.

\subsection{The Contribution of Track-best}

The objective does not necessarily decrease monotonically over restoration rounds, so track-best keeps the \emph{best drop set seen so far} instead of the last iterate. \cref{tab:supp_trackbest} measures its contribution at two round counts. At $5$ rounds it improves Chamfer from $0.8403$ to $0.7910$; at $10$ rounds it moves the result by only $0.002$. Five rounds with track-best already match ten without it ($0.7910$ against $0.7952$): given enough rounds the last iterate is itself a good set, and remembering the best one has little left to add. Track-best therefore does not raise the attainable quality; it reaches the same quality in fewer rounds.

\begin{table}[t]
\centering\small
\caption{\textbf{Track-best on and off.} DTU, VGGT, $22$ scans, highest operating point.
All four arms have the same measured sparsity, so the rows are directly comparable.}
\label{tab:supp_trackbest}
\begin{tabular}{llccc}
\toprule
rounds & track-best & sparsity & CD $\downarrow$ & AUC@$30^\circ$ $\uparrow$ \\
\midrule
$5$ & -- & $0.7274$ & $0.8403$ & $0.9939$ \\
$5$ & \checkmark & $0.7274$ & $\textbf{0.7910}$ & \textbf{$0.9958$} \\
\midrule
$10$ & -- & $0.7274$ & $0.7952$ & $0.9950$ \\
$10$ & \checkmark & $0.7274$ & $\textbf{0.7934}$ & \textbf{$0.9964$} \\
\bottomrule
\end{tabular}

\end{table}

\subsection{Reproducibility}

Running the same configuration twice on the same GPU gives \textbf{bit-identical}
results. On DTU with VGGT at the highest operating point, all four arms (dense,
SparseVGGT, HeSS, ReSS) gave $|\Delta \mathrm{CD}| = 0$ and $|\Delta \mathrm{AUC}| = 0$.
Nothing in the selection, the forward pass, or the scoring varies from run to run.
Every quality number in this paper was measured on a single GPU model (NVIDIA L40S); every latency number, on an RTX 4090.

\section{Dense Alignment for Precise Evaluation}
\label{sec:supp_dense_alignment}

DA3-Bench~\cite{lin2025depthanything3} aligns the predicted geometry to the ground truth with a Sim(3) transform, estimated by a Umeyama fit~\cite{umeyama1991umeyama} to the camera centers of the scene. This gives only as many samples as there are cameras ($49$ on DTU~\cite{aanaes2016dtu}): as sparsity grows and a few poses go badly wrong, those few dominate the fit, and the measured degradation mixes \emph{degraded geometry} with \emph{failed alignment}, changing the shape of the performance curve over sparsity.

We therefore estimate the transform from pixel-wise depth correspondences. Unprojecting the predicted and ground-truth depths, with validity masks from the backbone's own confidence threshold, makes each pixel its own correspondence, yielding tens of millions of pairs per scene and diluting the influence of a few wrong poses accordingly. We run the Umeyama fit on these pairs, trim the top quantile of residuals once, and refit, which keeps regions of collapsed geometry from dragging the global transform. The procedure is applied identically to every method and every sparsity level, and the same alignment enters the head importance calibration of HeSS (\cref{sec:supp_hess_impl}): the entire paper uses one definition of alignment.

\section{HeSS Implementation}
\label{sec:supp_hess_impl}

ReSS takes the per-head budgets $\{c_h\}$ as given and, for fairness of comparison, uses the budget allocation of HeSS~\cite{kim2026hess}. We recompute the allocation for all three backbones with a single procedure following the description in the HeSS paper; since HeSS and ReSS share the result, differences in allocation do not enter the comparison.

\paragraph{Procedure.}
HeSS measures the importance of each head by the trace of the Fisher information of that head's QKV weight gradients, averaged over a calibration set. Following the original procedure, we keep the CO3Dv2~\cite{reizenstein21co3d} dev split, $20$ views, and per-scene averaging, accumulate $\operatorname{tr}(g g^{\top}) = \lVert g \rVert^{2}$, and compute the loss under the project-wide pixel-wise alignment (\cref{sec:supp_dense_alignment}). The two losses (point cloud and camera position) each yield one table, combined as
\[
s = \lambda \cdot s_{\mathrm{cam}} + (1-\lambda) \cdot s_{\mathrm{pc}}.
\]
The head scores are turned into $\{c_h\}$ by proportional allocation with iterative clipping at the cap; the allocation is fixed after calibration and adds no inference-time cost.

\paragraph{$\lambda$ differs across backbones.}
Which loss carries head importance depends on the backbone. For VGGT and $\pi^3$ we use the values reported by HeSS~\cite{kim2026hess}: $\lambda = 0.5$ (Sec.~S4.1) and $\lambda = 0.0$ (Sec.~S4.5). HeSS reports no value for DepthAnything3, so we sweep $\lambda \in \{0.0, 0.5, 1.0\}$ as HeSS does for $\pi^3$ and take $\lambda = 1.0$, which gives the best DTU pose and ScanNet++ F1 (\cref{tab:supp_hess_sweep}). DTU Chamfer slightly favors $\lambda = 0.5$, but ReSS stays ahead of HeSS on both datasets at either value (\cref{tab:supp_da3}).

\begin{table}[t]\centering\small
\caption{\textbf{$\lambda$ sweep for DepthAnything3.} $22$ DTU scenes and $20$ ScanNet++ scenes at the highest compression operating point; measured sparsity $0.7963$ on DTU and $0.7946$ on ScanNet++, identical across the three settings. The best value is in bold.}
\label{tab:supp_hess_sweep}
\begin{tabular}{lccc|c}
\toprule
 & \multicolumn{3}{c|}{DTU} & ScanNet++ \\
$\lambda$ & CD $\downarrow$ & AUC@$30^\circ$ $\uparrow$ & AUC@$5^\circ$ $\uparrow$ & F1 $\uparrow$ \\
\midrule
$0.0$ & $2.3221$ & $0.9425$ & $0.7137$ & $0.7204$ \\
$0.5$ & $\mathbf{2.1541}$ & $0.9729$ & $0.8390$ & $0.7416$ \\
$1.0$ & $2.2054$ & $\mathbf{0.9774}$ & $\mathbf{0.8650}$ & $\mathbf{0.7551}$ \\
\bottomrule
\end{tabular}
\end{table}

\section{Fused Kernel Implementation}
\label{sec:supp_fused_kernel_impl}

One restoration round consists of computing the restore/drop gains \cref{eq:restore-delta,eq:drop-delta}, masking against the inside and outside of the drop set, exact top-$m$ selection for restores and for drops, and updating the drop set and the scalars $G$ and $K$. Each is a cheap operation over an $(n_h, q_{\mathrm{blk}}, k_{\mathrm{blk}})$ tensor, but each incurs its own round trip to memory, so the round is bound by bandwidth rather than computation. We therefore fuse all of the above into a single Triton kernel. Each query row is processed in registers in one sweep, from gain computation to the drop-set update, with no intermediate tensor written to memory and only the new drop mask written out. The GEMM behind the inner-product term of the gains stays outside the kernel, taking the cuBLAS output as is. The effect of fusion is reported in the w/o fused kernel row of \cref{tab:cost_analysis}.

The kernel produces bit-for-bit the same selection as the unfused path: the GEMM is not moved, so no accumulation order changes; the elementwise operations round at the same points as the unfused path; and the top-$m$ is exact, with no approximation. This identity holds within the supported configurations; outside them, the implementation falls back to the unfused path. The cap $k_{\mathrm{blk}} \le 4096$ comes from holding one row in registers, corresponds to about $253$ views, and is conservative; every experiment in this paper is within this range.

\section{Porting LLM Sparse Attention Baselines}
\label{sec:supp_llm_baselines}

\paragraph{Common conditions.}
The three methods replace only the mask-construction step and share everything else with our method: the same grid of query blocks of $128$ and key blocks of $64$, the same block-sparse kernel, the same sink rule that always keeps the trailing block holding the special tokens, the same floor of at least two key blocks per query row that applies to every method in the paper, and the same definition of measured sparsity. The always-keep rules that each method defines in its own paper are kept on top. All three set their budget through a threshold, so for each of our grid points we pre-measure, on a subset of scenes, the threshold whose measured sparsity comes closest, and fix it; the maximum residual from the target sparsity is $0.023$.

All three methods are designed for causal attention, whereas the global attention of 3D ViTs is bidirectional. We run all three without a causal mask and reinterpret every causality-dependent rule so that it keeps its intent under bidirectional attention. The antidiagonal score of XAttention is computed per block pair and does not depend on token order, and SpargeAttn calls the non-causal path already present in its official implementation. FlexPrefill has three causality-dependent rules, all listed in \cref{tab:llm_port_choices}.

\begin{figure}[t]
    \centering
    \includegraphics[width=\linewidth]{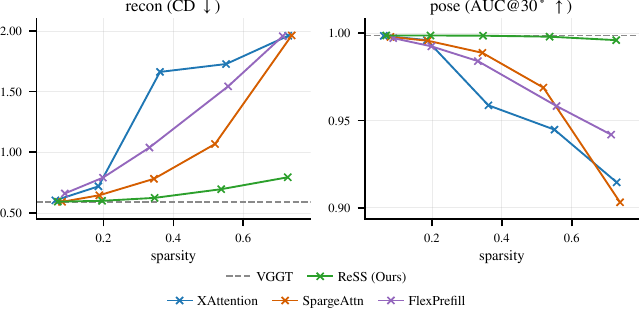}
    \vspace{-1.6em}
    \caption{\textbf{Comparison with sparse attention for LLMs.} DTU, VGGT~\cite{wang2025vggt} backbone.
    Dashed line denotes dense attention.
    \textcolor{mygreen}{ReSS} preserves dense performance better than the three methods ported from LLM sparse attention.}
    \label{fig:llm_baselines}
\end{figure}

\begin{figure*}[t]
\centering
\includegraphics[width=\linewidth,height=0.88\textheight,keepaspectratio]{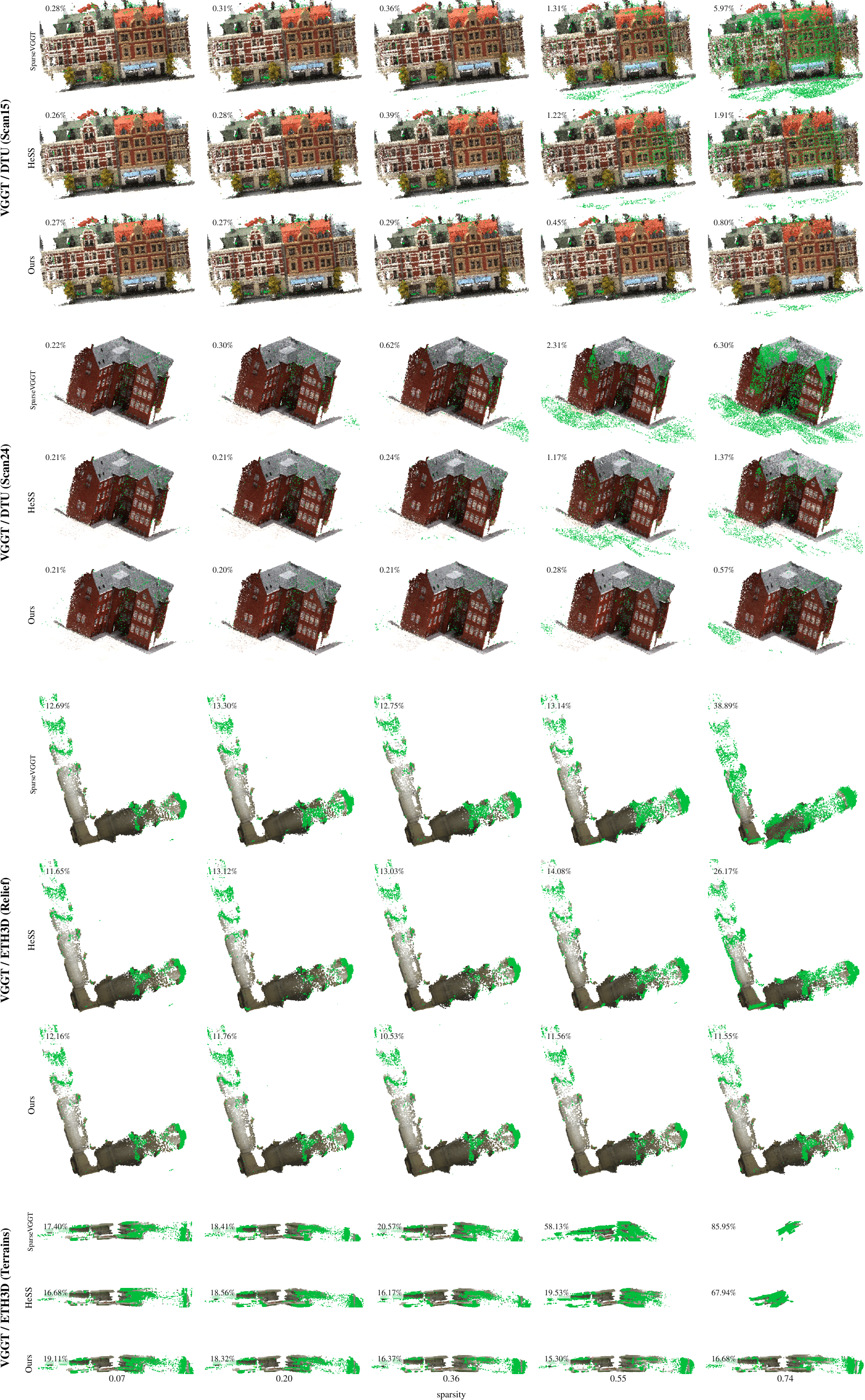}
\caption{\textbf{Qualitative results on DTU and ETH3D (VGGT).} Each block is one scene;
rows are the methods and columns the measured sparsity, matched across methods. A point is drawn in green when its distance to the ground truth exceeds a per-benchmark threshold (5\,mm on DTU, and the F1 threshold of 0.25\,m on ETH3D), and each panel reports the fraction of points above it.}
\label{fig:supp_qual1}
\end{figure*}

\begin{figure*}[t]
\centering
\includegraphics[width=\linewidth,height=0.95\textheight,keepaspectratio]{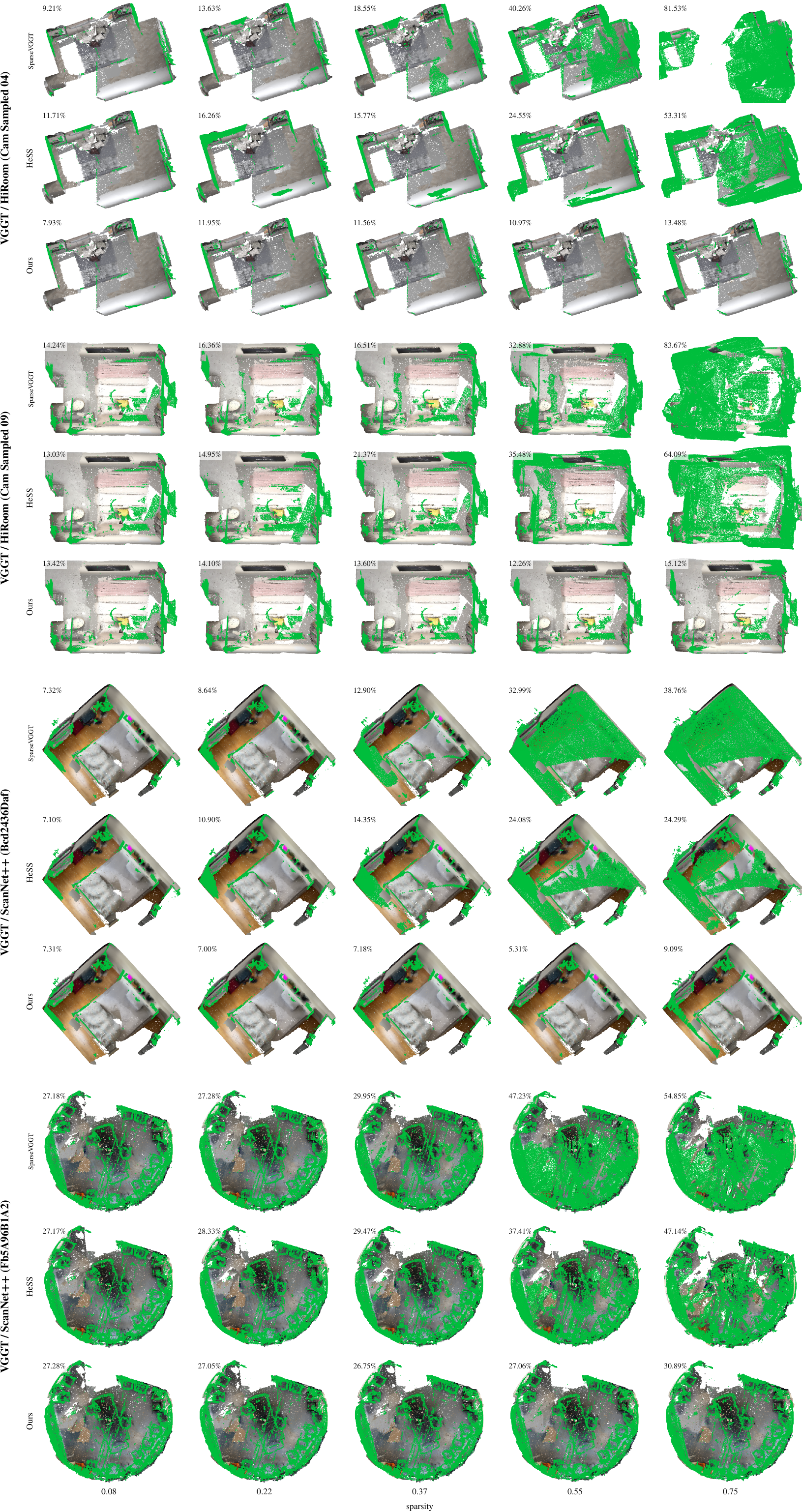}
\caption{\textbf{Qualitative results on HiRoom and ScanNet++ (VGGT).} Read as in
\cref{fig:supp_qual1}. The accuracy threshold is $0.05$\,m on both benchmarks.}
\label{fig:supp_qual2}
\end{figure*}

\paragraph{Porting decisions and their direction.}
All three methods are evaluated under a condition their papers do not cover: bidirectional sequences at the $10^5$-token scale. Transferred literally, two of the three either fail to compress at all or operate on biased statistics. \Cref{tab:llm_port_choices} lists every decision: each one favors the baseline, preserves the original rule's intent, or, in one case, is marked as having no established direction. The gaps in \cref{fig:llm_baselines} and \cref{tab:llm_baselines} are therefore unlikely to be artifacts of the port.

\begin{table*}[t]
    \centering
    \small
    \setlength{\tabcolsep}{4pt}
    \caption{\textbf{Porting decisions.} Each row is a point where the original rule fails on 3D ViTs, together with our treatment; the parenthetical states whom the treatment favors, where this can be established. The favorable decisions are cases where, without the treatment, the baseline either could not be evaluated at all or would be penalized as an artifact of the port.}
    \label{tab:llm_port_choices}
    \begin{tabular}{@{}p{0.19\linewidth}p{0.38\linewidth}p{0.38\linewidth}@{}}
        \toprule
        Rule in the original paper & Why it fails on 3D ViTs & Our treatment \\
        \midrule
        Causal attention
        & Global attention is bidirectional, with no query--key ordering constraint.
        & Run without a causal mask; order-dependent rules are reinterpreted to keep their intent (\emph{neutral}: preserves the rule's intent). \\
        \addlinespace
        SpargeAttn~\cite{zhang2025spargeattn} similarity gate $\tau_\text{sim}=0.6$
        & Nearly every VGGT block is judged self-dissimilar and reverted to dense: measured sparsity $0.006$ at the default cumulative threshold of the official implementation ($0.98$), and only $0.062$ even at $0.1$.
        & Disable the gate and sweep the cumulative threshold alone (\emph{favors the baseline}: no compression curve exists otherwise). Without the gate, the selection effectively reduces to the SparseVGGT rule of a cumulative threshold on pooled query/key inner products. \\
        \addlinespace
        FlexPrefill~\cite{lai2025flexprefill} representative queries $=$ last $128$
        & In a causal LM those queries are the only ones that have seen the full context; under bidirectional attention they are a biased sample covering part of the last frame.
        & Sample uniformly across the sequence (\emph{favors the baseline}: the literal rule gives a worse estimate). \\
        \addlinespace
        FlexPrefill~\cite{lai2025flexprefill} slash lines $=$ constant offsets $q-k$
        & In a causal LM only non-negative offsets exist, so the slash family is the lower triangle; under bidirectional attention both signs carry mass, and searching the causal range alone would discard half of the candidate lines.
        & Search the full signed range $-(T_k-1)\ldots(T_q-1)$ (\emph{neutral}: preserves the rule's intent). \\
        \addlinespace
        FlexPrefill~\cite{lai2025flexprefill} always-keep $=$ first/last key block of each query block
        & In a causal LM the last key block is the diagonal block; under bidirectional attention it merely points at the end of the sequence, and the rule loses its intent.
        & Keep the first block and the diagonal block, i.e., the blocks the rule pointed at in a causal LM (\emph{neutral}: preserves the rule's intent). \\
        \addlinespace
        FlexPrefill~\cite{lai2025flexprefill} vertical/slash line granularity
        & The paper's Algorithm 3 defines the lines at token resolution, whereas the released implementation sum-pools the vertical and slash scores into blocks of $128$ before selecting; neither granularity is the $64$-wide key block of the shared kernel.
        & Select at token resolution, following the paper, and keep a key block if a selected line passes through it (\emph{direction not established}: rasterizing adds blocks relative to a token-exact mask, whereas the released implementation keeps coarser, wider blocks). \\
        \addlinespace
        XAttention~\cite{xu2025xattention} threshold $\tau$
        & At $10^5$ tokens the paper's $\tau=0.9$ already corresponds to sparsity $0.39$, so values near it never reach the low-compression regime.
        & Extend the sweep up to $\tau=0.9999$ to cover our full grid (\emph{favors the baseline}: with the original range, the low-compression points are empty and no comparison is possible). \\
        \bottomrule
    \end{tabular}
\end{table*}

\paragraph{Observation.} 
XAttention and FlexPrefill share one property: each sizes its mask by a threshold on attention probability mass alone, per query-block row for XAttention and per head over the whole block map for FlexPrefill, so beyond the common two-block floor, no row is guaranteed a budget. SparseVGGT instead fixes the count per row through its top-$k$ rule. On one scene, we measure the attention mass actually captured by masks of equal measured sparsity. XAttention matches SparseVGGT on average ($0.820$ vs.\ $0.819$) but falls to $0.655$ vs.\ $0.753$ on the bottom $5\%$ of rows, so the loss concentrates on a few rows instead of spreading across queries.

\begin{table}[t]
    \centering
    \small
    \setlength{\tabcolsep}{2pt}
    \caption{\textbf{Comparison with sparse attention for LLMs.} DTU, VGGT~\cite{wang2025vggt} backbone, 22 scans, at two matched sparsity levels. The last block lists three methods proposed for LLM long-context prefill, ported onto the same block grid and the same sparse kernel as ReSS, so that only the block selection criterion differs.}
    \label{tab:llm_baselines}
    \begin{tabular*}{\linewidth}{@{\extracolsep{\fill}}lcccc@{}}
        \toprule
        & \multicolumn{2}{c}{Chamfer $\downarrow$} & \multicolumn{2}{c}{Pose AUC@$30^\circ$ $\uparrow$} \\
        \cmidrule(lr){2-3} \cmidrule(lr){4-5}
        Sparsity & $0.35$ & $0.73$ & $0.35$ & $0.73$ \\
        \midrule
        VGGT (dense)                                 & 0.588 & 0.588 & 0.999 & 0.999 \\
        \midrule
        SparseVGGT~\cite{wang2026sparseVGGT}         & 0.727 & 1.338 & 0.991 & 0.963 \\
        HeSS~\cite{kim2026hess}                      & 0.718 & 0.982 & 0.995 & 0.982 \\
        \midrule
        XAttention~\cite{xu2025xattention}           & 1.663 & 1.961 & 0.959 & 0.915 \\
        SpargeAttn~\cite{zhang2025spargeattn}        & 0.781 & 1.964 & 0.989 & 0.903 \\
        FlexPrefill~\cite{lai2025flexprefill}        & 1.039 & 1.957 & 0.984 & 0.942 \\
        \textbf{ReSS (Ours)}                         & \textbf{0.623} & \textbf{0.791} & \textbf{0.999} & \textbf{0.996} \\
        \bottomrule
    \end{tabular*}
\end{table}

\section{Porting Video Sparse Attention Baselines}
\label{sec:supp_svg_baselines}

\paragraph{Porting strategy.}
Unlike the LLM family, the cost of this family lies in its variable block layouts and their mask construction, so we use the upstream kernel and harness exactly as released. The dense-step schedule has no counterpart in a single forward pass; the k-means warm start and the permutation are all kept as is, hyperparameters are the upstream defaults without retuning, and the only knob we sweep is the one each method exposes in the upstream scripts ($\texttt{--sparsity}$ for SVG, $\texttt{--top\_p\_kmeans}$ for SVG2 and SVG-EAR). The one exception is the protected tokens: the block holding the camera/register tokens is placed in the upstream slot for context tokens and always kept, the same rule as for every other baseline in the paper.

\paragraph{Choice of SVG variant.}
The upstream implementation ships a separate mask definition per model family; of these, the Wan and HunyuanVideo variants are the candidates, and we use the latter. All variants share one formula that converts a target density into band widths after subtracting the cost of the always-kept context tokens; the formula is derived for the HunyuanVideo mask shape, and the Wan family, which has no context tokens, shares it only because that term vanishes. Measured elementwise on the VGGT/ScanNet++ shapes, the HunyuanVideo variant also hits the target density more accurately.

\begin{table}[t]\centering\small
\caption{\textbf{Latency against the k-means iteration count.} VGGT, $100$ views, RTX 4090. The SVG2 row is a linear fit over the sweep; dense and SparseVGGT do not depend on the iteration count, which confirms that the sweep touches nothing but k-means.}
\label{tab:supp_svg_kmeans}
\begin{tabular}{lcc}
\toprule
 & $t$ (s) & at $50$ iter \\
\midrule
dense       & $10.47$ (constant)                   & $10.47$ \\
SparseVGGT  & $7.58$ (constant)                    & $7.58$ \\
SVG2        & $12.40 + 0.199 \cdot \mathrm{iter}$  & $22.35$ \\
\bottomrule
\end{tabular}
\end{table}

\paragraph{The k-means cost does not amortize.}
The upstream scripts ($\texttt{--zero\_step\_kmeans\_init}$) run $50$ iterations of k-means at the first denoising step of each layer, which is still a dense step, and $2$ warm-started iterations at each of the remaining $49$ steps, dense or sparse, an average of $2.96$ iterations per step. VGGT has a single forward pass, so every global layer pays the full $50$ iterations of cold k-means. The speedup of this family presupposes centroid reuse across denoising steps, and a single-pass reconstructor has no denominator to divide by. To confirm this, we sweep the iteration count directly and measure latency (\cref{tab:supp_svg_kmeans}): the cost is exactly linear in the iteration count. The point is the intercept: applying the upstream amortized rate ($2.96$ iterations) gives $13.0$\,s, and even with k-means entirely free the latency is $12.40$\,s, both above the dense $10.47$\,s. The cost does not sit in k-means alone but in re-planning the variable block layout at every layer, and no adjustment of the iteration count closes this gap.

\end{document}